\documentclass{article}

\PassOptionsToPackage{numbers, compress}{natbib}

\author{
  David I. Inouye \quad\quad Pradeep Ravikumar \quad\quad Inderjit S. Dhillon \\
  {Dept. of Computer Science, University of Texas at Austin, Austin, TX}\\
  \texttt{\{dinouye,pradeepr,inderjit\}@cs.utexas.edu} \\
}
\date{}

\title{Generalized Root Models: Beyond Pairwise Graphical Models \\ for Univariate Exponential Families}

\usepackage[utf8]{inputenc} 
\usepackage[T1]{fontenc}    
\usepackage{hyperref}       
\usepackage{url}            
\usepackage{booktabs}       
\usepackage{amsfonts}       
\usepackage{nicefrac}       
\usepackage{microtype}      
\usepackage{natbib}

\usepackage{enumitem}
\setitemize{noitemsep,topsep=0pt,parsep=0pt,partopsep=0pt}
\setenumerate{noitemsep,topsep=0pt,parsep=0pt,partopsep=0pt}
\usepackage{times}
\usepackage{subfigure} 
\usepackage{graphicx}
\usepackage{mathtools} 
\usepackage{amsthm, amssymb, amscd, amsfonts} 
\usepackage[ampersand]{easylist} 
\usepackage{multirow} 
\usepackage{booktabs} 
\usepackage{bm} 
\usepackage{dcolumn} 
\usepackage{wrapfig}
\usepackage{bbm}
\usepackage{verbatim}
\usepackage{fullpage}

\usepackage{macros}

\begin{document} \maketitle
\begin{abstract}
We present a novel $\k$-way high-dimensional graphical model called the Generalized Root Model (GRM) that explicitly models dependencies between variable sets of size $\k \geq 2$---where $\k=2$ is the standard pairwise graphical model.  This model is based on taking the $\k$-th root of the original sufficient statistics of any univariate exponential family with positive sufficient statistics, including the Poisson and exponential distributions.  As in the recent work with square root graphical (SQR) models \cite{Inouye2016}---which was restricted to pairwise dependencies---we give the conditions of the parameters that are needed for normalization using the radial conditionals similar to the pairwise case \cite{Inouye2016}.  In particular, we show that the Poisson GRM has no restrictions on the parameters and the exponential GRM only has a restriction akin to negative definiteness.  We develop a simple but general learning algorithm based on $\ell_1$-regularized node-wise regressions.  We also present a general way of numerically approximating the log partition function and associated derivatives of the GRM univariate node conditionals---in contrast to \cite{Inouye2016} which only provided algorithm for estimating the exponential SQR. To illustrate GRM, we model word counts with a Poisson GRM and show the associated $\k$-sized variable sets. We finish by discussing methods for reducing the parameter space in various situations.
\end{abstract}

\section{Introduction}
Most standard graphical models are restricted to pairwise dependencies between variables.  For example, the Ising model for binary data and the multivariate Gaussian for real-valued data are popular pairwise graphical models.  However, real-world data often exhibits triple-wise, or more generally $\k$-wise dependencies.  For example, the words \emph{deep}, \emph{neural} and \emph{network} often occur together in recent research papers---note that this \emph{triple} of words refers to something more specific than any of the two words without the third word, i.e. if a document only contains \emph{neural} and \emph{network} but not \emph{deep}, then this may be a more classical paper about shallow neural networks.  In the biological domain, genetic, metabolic and protein pathways play an important role in studying the development of diseases and possible interventions.  These pathways are known to be complex and involve many genes or proteins rather than just simple pairwise interactions.\footnote{\url{https://www.genome.gov/27530687/}}

Thus, we seek to begin bridging this gap between pairwise models and complex real-world data that contain complex $\k$-wise interactions by defining a class of $\k$-wise graphical models called Generalized Root Models (GRM), which can be instantiated for any $\k \geq 1$ and any univariate exponential family with positive sufficient statistics including the Gaussian (using the $\x^2$ sufficient statistic), Poisson and exponential distributions.  We estimate the graphical model structure and parameters using $\ell_1$-regularized node-wise regressions similar to previous work \cite{Ravikumar2010,Yang2015,Inouye2015,Inouye2016}.  However, unlike previous work, because the log partition function of the GRM node conditionals is not known in closed-form---even for the previous work considering the pairwise case\cite{Inouye2016}---we develop a novel numerical approximation method for the GRM log partition function and related derivatives.  In addition, we present a Newton-like optimization algorithm similar to \cite{Hsieh2011} to solve the node-regressions---which significantly reduces the number of numerical log partition function approximations needed compared to gradient descent.  Finally, we demonstrate the GRM model and parameter estimation algorithm on real-world text data.

\section{Related Work}
This paper generalizes the square root graphical model (SQR) from \cite{Inouye2016}, which only considers pairwise dependencies.  \cite{Inouye2016} followed the idea of constructing a joint distribution by defining the form of the node-conditional distributions as in \cite{Yang2015} but introduced the idea of taking the square root of the sufficient statistics $\T(\x)$ to form a pairwise term $\sqrt{\T(\x_\vi)}\sqrt{\T(\x_\vit)}$ which is linear $O(\T(\x))$ rather than the pairwise term $\T(\x_\vi)\T(\x_\vit)$ in \cite{Yang2015} which is quadratic $O(\T(\x)^2)$.  This elegant modification allowed for arbitrary \emph{positive} and \emph{negative} dependencies in the Poisson SQR graphical model whereas the Poisson graphical model in \cite{Yang2015} only permitted negative dependencies---a crucial limitation of the Poisson models from \cite{Yang2015}.  While \cite{Yang2013} proposed three modifications to the original Poisson models as defined in \cite{Yang2015}, the modifications lead to distributions with either Gaussian-esque thin tails or truncated distributions which required unintuitive cutoff points where the probability mas may concentrate near the corners of the distribution \cite{Yang2013}.  Though SQR models have great promise, SQR models are limited to pairwise dependencies, and \cite{Inouye2016} did not provide an estimation algorithm for the Poisson SQR model because the node conditional log partition function is not known in closed form.  Thus, this paper extends the SQR model class to include $\k$-wise interactions where $\k > 2$ and, in addition, instantiates a concrete approximation algorithm for the node conditional log partition function and associated derivatives.

In a somewhat different direction, latent variable models provide an implicit and indirect way of modeling complex dependencies.  Generally, though the explicit dependencies in latent variable models are only pairwise, many variables can be related implicitly through a latent variable.  For example, mixture models associate a discrete latent variable with every instance which implicitly introduces dependencies.  Other more complex latent variable models such as topic models \cite{Blei2012,Blei2006} can introduce even more implicit dependencies in interesting ways.  While latent variable models have proven to be practically effective in helping to model complex dependencies, the development of GRM models in this paper is distinctive and somewhat orthogonal to latent variable models.  As opposed to implicitly modeling dependencies through latent variables, the GRM model explicitly models dependencies between observed variables.  Thus, the discovered dependencies have an intuitive and obvious explanation in terms of the observed data variables.  In addition, GRM models can be seen as complementary to latent variable models because GRM models can be used as base distributions for these latent variable models.  For example, \cite{Inouye2014,Inouye2015} explore using count-valued graphical models in mixtures and topic models.  Thus, GRMs can provide new components from which to build more interesting models for real-world situations.  Finally, node-conditional models such as GRM can be estimated using convex optimization problems, which often have theoretical guarantees \cite{Ravikumar2010,Yang2015} whereas latent variable models often require optimizing a non-convex function and struggle with theoretical guarantees.



\paragraph{Notation}
Let $\vmax$ and $\imax$ be the number of dimensions and data instances respectively. Let $\Rp$ denote the set of nonnegative real numbers and $\Zp$ denote the set of nonnegative integers. Unless indicated otherwise, we denote vectors with boldface lower case letters (e.g. $\xvec$, $\nodepvec$) and their corresponding scalar values as normal lower case letters (e.g. $\x_\vi$, $\nodep_\vi$).  We denote the standard basis vectors as $\evec_\vi = [0,\cdots,0,1,0,\cdots,0]^T$ and the ones vector as $\evec = [1,1,\cdots,1]^T$.  Let $\bm{x}^p$ and $\sqrt[\j]{\bm{x}}$ to be the entry-wise power and $\j$-th root of the vector $\bm{x}$.
We denote tensors (or multidimensional arrays) with parenthesized superscripts as $X^{(\k)}$ where $\k$ is the order of the tensor.  For example, $A^{(2)} \in \R^{\vmax \times \vmax}$ is a matrix, $A^{(3)} \in \R^{\vmax \times \vmax \times \vmax}$ is a three dimensional tensor, and $A^{(\k)} \in \R^{\vmax \times^\k}$ is a $\k$-th order tensor.  We index tensors using brackets and subscripts, e.g. $[A^{(3)}]_{1,2,3}$ is a scalar value in the multidimensional array at index $(1,2,3)$. We define $[A^{(\ell)}]_{\vi} \in \R^{\vmax \times^{\ell-1}}$ to be a sub tensor created by fixing the last index to $\vi$ and letting the others vary---in MATLAB colon indexing notation, this would be $A(:,:,\dots,:,\textrm{\vi})$.  For example, if $A^{(3)} \in \R^{\vmax \times^3}$, then $[A^{(3)}]_{\vi} \in \R^{\vmax \times \vmax}$ is a matrix corresponding to the $\vi$-th slice of the tensor $A^{(3)}$.
We define $\circ$ to be the outer product operation.  For example, $\xvec \circ \xvec = \xvec \xvec^T \in \R^{\vmax \times \vmax}$ and $\xvec \circ \xvec \circ \xvec \in \R^{\vmax \times^3}$, where $[\xvec \circ \xvec \circ \xvec]_{\vi_1\vi_2\vi_3} = \x_{\vi_1}\x_{\vi_2}\x_{\vi_3}$.  For more general sizes, we denote a $\k$-th outer product to be $\xvec \,\circ^{\k} = \xvec \circ \cdots \circ \xvec$ such that there are $\k$ copies of $\xvec$ and the result is a $\k$-th order tensor.  We define $\xvec \,\circ^0 = \evec = [1,1,\cdots,1]^T$. We also denote the inner product operation of two tensors as $\left\langle A^{(\k)}, B^{(\k)}  \right\rangle = \sum_{\vi_1,\cdots,\vi_\k} a_{\vi_1,\cdots,\vi_\k} b_{\vi_1,\cdots,\vi_\k}$.

\section{Generalized Root Model}
With the notation given in the previous section, we will define the GRM model.  First, let the sufficient statistic and log base measure of a univariate exponential family be denoted as $\T(\x)$ and $\B(\x)$ respectively. We will also define the domain (or support) of the random variable to be $\domain$ and it's corresponding measure to be $\mu(\x)$, which is either the counting measure or Lebesgue measure depending on whether $\x$ is discrete or continuous. 

Let us denote a new $\j$-th root sufficient statistic $\widetilde{\T}_\j(\x) = \sqrt[j]{\T(\x)}$ except in the case when $\T(\x) = f(\x)^{c\j}$ where $c$ is an even positive integer.  If $\T(\x) = f(\x)^{c\j}$, then we simplify $\widetilde{\T}_\j(\x) \equiv f(\x)^{c}$ (rather than the usual $|f(\x)|^c$). For example, if $\T(\x) = \x^2$, then $\widetilde{\T}_2(\x) \equiv x$ (rather than $|x|$). As in \cite{Inouye2016}, this nuanced definition is necessary to recover the multivariate Gaussian distribution.  However, for notational simplicity, we will merely write $\sqrt[\j]{\x}$ for $\widetilde{\T}_\j(\x)$ throughout the paper.  Note that $\widetilde{\T}_\j(\x) = \sqrt[\j]{\x}$ for the Poisson and exponential GRM models.  Using this simplified notation, we can define the Generalized Root Model for $\k \leq \vmax$ as:
\begin{align}
\Pr( \instvec \given \multipall ) &= \exp\left( \sum_{\j=1}^\k \sum_{\ell=1}^\j \left\langle \multip^{(\ell)}_{(\j)}, \sqrt[\j]{\xvec} \, \circ^{\ell} \right\rangle + \textstyle{\sum_{\vi}} \B(\x_\vi) - \A(\multipall) \right) \label{eqn:full-model} \\
\A(\multipall) &= \log \int_\domain \exp\left( \sum_{\j=1}^\k \sum_{\ell=1}^\j \left\langle \multip^{(\ell)}_{(\j)}, \sqrt[\j]{\xvec} \, \circ^{\ell} \right\rangle + \textstyle{\sum_{\vi}} \B(\x_\vi) \right) \mathrm{d}\mu(\xvec)
\, , \label{eqn:full-model-A}
\end{align}
where $\A(\multipall)$ is the joint log partition function, $\multipall = \left\{\multip_{(\j)}^{(\ell)}: \j \in \{1,\cdots,\k\}, \ell \leq \j \right\}$, $\multip^{(\ell)}_{(\j)}$ are super symmetric tensors of order $\ell$ which are zero whenever two indices are the same. More formally, letting $\pi(\cdot)$ be an index permutation:
\begin{align}
\multip^{(\ell)}_{(\j)} \in \left\{ A^{(\ell)} \,:
\begin{array}{lll}
\left[A^{(\ell)}\right]_{\vi_1,\cdots,\vi_\ell} &= [A^{(\ell)}]_{\pi(\vi_1,\cdots,\vi_\ell)} & \forall \pi(\cdot), \\
\left[A^{(\ell)}\right]_{\pi(\vi_u,\vi_v,\cdots,\vi_\ell)} &=0 & \forall \{(u, v, \pi(\cdot)): u \neq v, \vi_u = \vi_v \}
\end{array}
\right\}
\, .
\label{eqn:parameter-constraints}
\end{align}
Note that the non-zeros of $\multip^{(\ell)}_{(\j)}$ define $\ell$-sized variable sets (or cliques) of the underlying graphical model.

\subsection{Special Cases}
We now consider several special cases of this model to build some understanding of the GRMs connection to previous models.  The independent model is trivially recovered if $\k = 1$: $\Pr(\xvec \given \multip^{(1)}_{(1)}) = \exp\left( \langle \multip^{(1)}_{(1)}, \xvec \rangle + \textstyle{\sum_{\vi}} \B(\x_\vi) - \A( \multipall ) \right)$.

\paragraph{Square Root Graphical Model \cite{Inouye2016}}
Another special case is the previous SQR models (i.e. $\k=2$) from \cite{Inouye2016} by taking (using the notation from \cite{Inouye2016}) $\multip^{(1)}_{(1)} = \text{diag}(\edgepmat)$, $\multip^{(1)}_{(2)} = \nodepvec$ and $\multip^{(2)}_{(2)} = \tilde{\edgepmat}$, where $\text{diag}(\edgepmat)$ is a column vector of the diagonal entries and $\tilde{\edgepmat}$ has the same off-diagonal entries as $\edgepmat$ but is zero along the diagonal.  Thus, the SQR model can be written as:
\begin{align*}
\Pr(\xvec \given \multip^{(1)}_{(1)}, \multip^{(1)}_{(2)}, \multip^{(2)}_{(2)}) = \exp\left( \langle \multip^{(1)}_{(1)}, \xvec \rangle + \langle \multip^{(1)}_{(2)}, \sqrt[2]{\xvec} \rangle + \langle \multip^{(2)}_{(2)}, \sqrt[2]{\xvec} \circ \sqrt[2]{\xvec} \rangle + \textstyle{\sum_{\vi}} \B(\x_\vi) - \A( \multipall ) \right) \, .
\end{align*}

\paragraph{Simplified Model with Only Strongest Interaction Terms}
We consider another special case such that only the strongest interaction (i.e. when $\ell = \j$) terms are non-zero:
\begin{align}
\Pr(\xvec \given \multipall) = \exp\left( \sum_{\j=1}^\k \langle \multip^{(\j)}_{(\j)}, \sqrt[\j]{\xvec} \,\circ^\j \rangle  \,+ \, \textstyle{\sum_{\vi}} \B(\x_\vi) - \A( \multipall ) \right) \, .
\label{eqn:simplified-model}
\end{align}
This restricted parameter space forces $\j$-wise dependencies to only be through the $\j$-th root term.  For example, pairwise interactions are only available through the sufficient statistic $\sqrt[2]{\x_\vi\x_\vit}$ and ternary interactions are only available through the sufficient statistic $\sqrt[3]{\x_\vi\x_\vit\x_\vir}$.  Without this restriction interactions would be allowed through multiple terms, e.g. pairwise interactions would be allowed through multiple sufficient statistics $\sqrt[2]{\x_\vi\x_\vit}, \sqrt[3]{\x_\vi\x_\vit},\cdots,\sqrt[\k]{\x_\vi\x_\vit}$. Thus, this simplified model is more interpretable and easier to learn while still retaining the strongest $\j$-wise interaction terms.  For our experiments, we assume this simplified model unless specified otherwise.

\subsection{Conditional Distributions}
\begin{figure}[!ht]
\centering
\includegraphics[width=0.7\columnwidth]{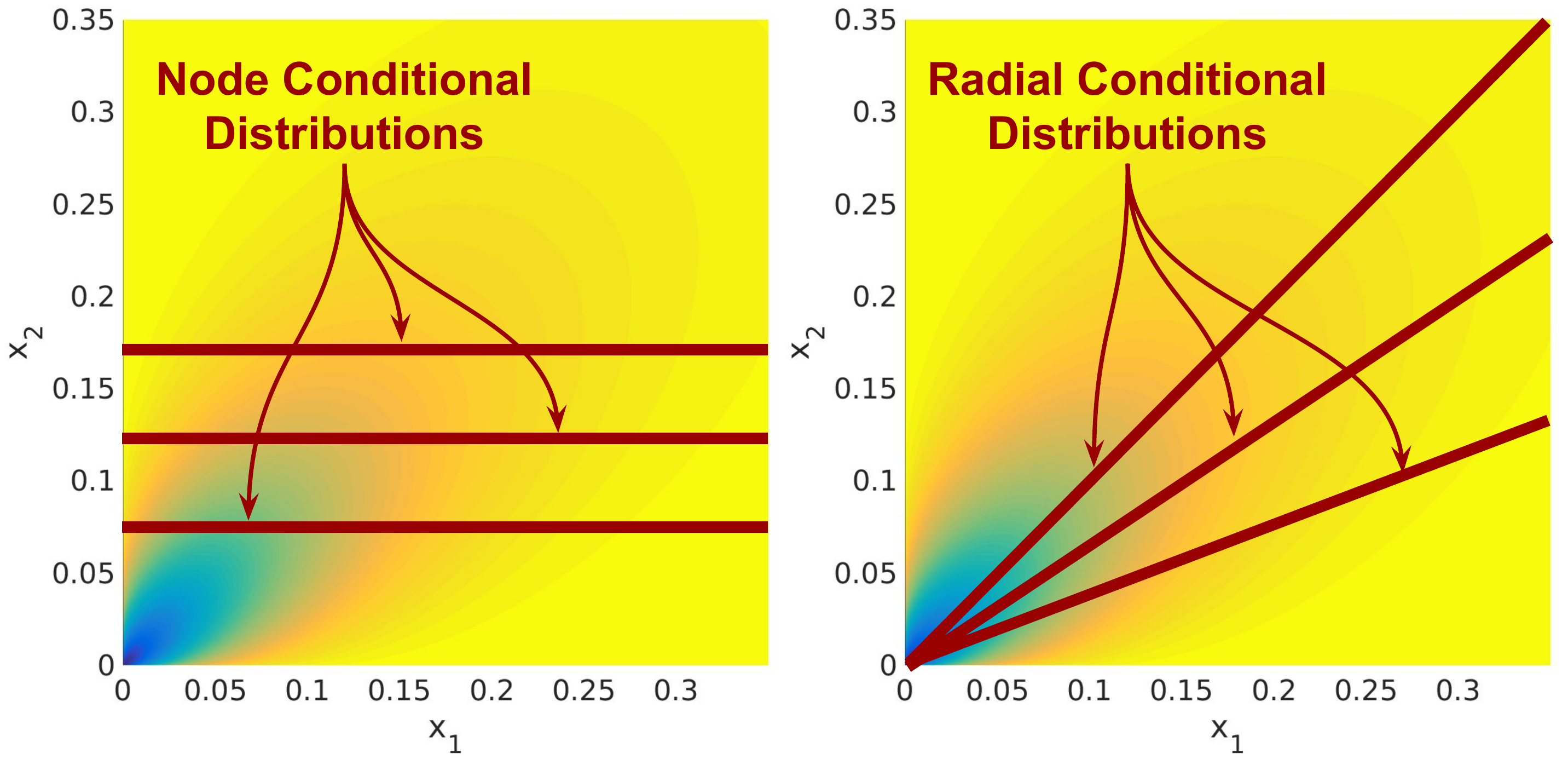}
\vspace{-1.8em}
\caption{\emph{Node} conditional distributions (left) are univariate probability distributions of one variable assuming the other variables are given while \emph{radial} conditional distributions are univariate probability distributions of vector scaling assuming the vector direction is given.  Both conditional distributions are helpful in understanding SQR graphical models. (Figure reproduced from \cite{Inouye2016} with permission.)}
\label{fig:conditional-illustration}
\end{figure}

As in \cite{Inouye2016}, we derive both the \emph{node} conditionals and the \emph{radial} conditional distributions.  An illustration of these two types of univariate conditional distributions can be seen in Fig.~\ref{fig:conditional-illustration}.  This \emph{node} conditional distribution is critical for the parameter estimation that will be described in later sections; whereas the \emph{radial} conditional distributions are critical for showing the normalization of GRM models.

\subsubsection{Node Conditionals}
\label{sec:node-conditionals}
The node conditionals are as follows (see appendix for full derivation):
\begin{align}
\Pr(\x_\vi \given \xvec_{-\vi}, \multipall) &\propto \exp\left( \sum_{\j=1}^\k \natp_{\j\vi} \x_\vi^{1/\j} + \B(\x_\vi) \right)
\, ,
\end{align}
where $\xvec_{-\vi}$ is all other variables except $\x_\vi$, $\natp_{\j\vi} = \sum_{\ell=1}^\j  \left\langle \left[\multip^{(\ell)}_{(\j)}\right]_{\vi}, \ell \sqrt[\j]{\xvec} \,\circ^{\ell-1} \right\rangle$.  This is a univariate exponential family with sufficient statistics $\x_\vi^{1/\j}$, natural parameters $\natp_{\j\vi}$ and base measure $\B(\x_\vi)$.  Note that this reduces to the original exponential family if the interaction terms $\natp_{2\vi}=\cdots=\natp_{\k\vi} = 0$.

\subsubsection{Radial Conditionals}
\label{sec:radial-conditionals}
As in \cite{Inouye2016}, we define the \emph{radial} conditional distribution by fixing the unit direction $\uvec = \frac{\xvec}{\|\xvec\|_1}$ of the sufficient statistics but allowing the scaling $\radx = \|\xvec\|_1$ to be unknown.  Thus, we get the following \emph{radial} conditional distribution (see appendix for derivation):
\begin{align}
\Pr( \instvec = \radx \uvec \given \uvec, \multipall ) &\propto \exp\left( \sum_{r\in\mathcal{R}} \natp_r(\uvec) \radx^r + \tilde{\B}_{\uvec}(\radx) \right)
\, ,
\end{align}
where $\mathcal{R} = \{\ell/\j : \j \in \{1,\cdots,\k\}, \ell \leq \j\}$ is the set of possible ratios, $\natp_r(\uvec) = \sum_{\{(\ell,\j): \ell/\j = r\} } \langle \multip^{(\ell)}_{(\j)}, \sqrt[\j]{\uvec} \, \circ^{\ell} \rangle$ are the exponential family parameters, $\radx^r$ are the corresponding sufficient statistics and $\tilde{\B}_{\uvec}(\radx) = \textstyle{\sum_{\vi}} \B(\radx\usca_\vi)$ is the base measure.  Thus, the radial conditional distribution is a univariate exponential family (as in \cite{Inouye2016}).  

\subsection{Normalization}
The previous exponential and Poisson graphical models \citep{Besag1974,Yang2015} could only model negative dependencies.  However, we generalize the results from the pairwise SQR model in \cite{Inouye2016} and show that GRM normalization for any $\k$ puts little to no restriction on the value of the parameters---thus allowing both positive and negative dependencies.  For our derivations, let $\Uset = \{ \uvec : \|\uvec\|_1 = 1, \uvec \in \Rp^\vmax\}$ be the set of unit vectors in the positive orthant. The GRM log partition function $\A(\multipall)$ can be decomposed into nested integrals over the unit direction and over the scaling $\radx$:
\begin{align}
\A(\nodepvec,\pmat) &= \log \!\! \int\limits_\Uset \!\! \int\limits_{\radset(\uvec)} \!\!\!\! \exp\left( \sum_{\j=1}^\k \sum_{\ell=1}^\j \left\langle \multip^{(\ell)}_{(\j)}, \sqrt[\j]{\radx\uvec} \, \circ^{\ell} \right\rangle + \textstyle{\sum_{\vi}} \B(\radx\usca_\vi) \right) \mathrm{d}\measure(\radx) \, \mathrm{d}\uvec \notag \\
&= \log \!\! \int\limits_\Uset \!\! \int\limits_{\radset(\uvec)} \!\!\!\! \exp\left( \sum_{r\in\mathcal{R}} \natp_r(\uvec) \radx^r + \tilde{\B}_{\uvec}(\radx) \right) \mathrm{d}\measure(\radx) \, \mathrm{d}\uvec
\end{align}
where $\radset(\uvec) = {\{\radx \in \Rp:\radx\uvec \in \domain\}}$, and $\measure$ and $\domain$ are the measure and domain (or support) of the random variable.  Because $\Uset$ is bounded, the joint distribution will be normalizable if the radial conditional distribution is normalizable---generalizing the results from \cite{Inouye2016} for $\k > 2$.  Informally, the radial conditional distribution converges if the asymptotically largest term of $\{\natp_r(\uvec) \radx^r\}\cup\{\B(\radx\usca_\vi)\}$ is monotonically decreasing at least linearly.\footnote{For more formal proofs, we refer the reader to \cite{Inouye2016}.}  We give several examples in the following paragraphs.

\paragraph{Gaussian GRM}
For the Gaussian GRM, we take the Gaussian univariate distribution with sufficient statistic $\T(\x) = \x^2$ and $\B(\x) = 0$.  When $\k = 2$ (i.e. the standard multivariate Gaussian), the largest radial conditional term is $\natp_1\x^2$ where $\natp_1 = \langle \multip^{(1)}{(1)}, \uvec^2 \rangle + \langle \multip^{(2)}{(2)}, \uvec \circ \uvec \rangle$.  Note that the radial conditional (i.e. a univariate Gaussian) is normalizable only if $\natp_1 < 0$ for all $\uvec \in \Uset$, which is equivalent to the positive definite condition on the Gaussian inverse covariance matrix. We can also consider a Gaussian-like model with $\k = 3$.  In this case, we have that $\natp_1 = \langle \multip^{(1)}_{(1)}, \uvec^2 \rangle + \langle \multip^{(2)}_{(2)}, \uvec \circ \uvec \rangle + \langle \multip^{(2)}_{(2)}, \uvec^{\frac{2}{3}} \circ \uvec^{\frac{2}{3}} \circ \uvec^{\frac{2}{3}} \rangle$ and we need $\natp_1 < 0\,\,\forall \uvec \in \Uset$.  Note that the Gaussian GRM models for $\k > 2$ are novel models to the authors' best knowledge.

\paragraph{Exponential GRM}
Because the exponential distribution also has a constant base measure like the Gaussian, the asymptotically largest term is $\natp_1\x$ and thus we must have that $\natp_1 < 0\,\,\forall \uvec \in \Uset$.  However, unlike the Gaussian, in the case of the exponential distribution $\Uset$ is only positive $\ell_1$-normalized vectors.  This is a significantly weaker condition on the parameters than for a Gaussian and allows strong positive and negative dependencies.

\paragraph{Poisson GRM}
For the Poisson distribution, the base measure is the asymptotically largest term $O(-\radx \log(\radx))$. Thus, as in \cite{Inouye2016}, the parameters can be arbitrarily positive or negative because eventually the base measure will ensure normalizability.  Note that this is true for arbitrarily large $\k$. 

\section{Parameter Estimation}
As in \cite{Yang2015,Inouye2015,Inouye2016}, we solve a set of independent $\ell_1$-regularized node-wise regressions for each node---based on the node conditional distributions in Sec.~\ref{sec:node-conditionals}---using a Newton-like method for convex optimization with an non-smooth $\ell_1$ penalty as in \cite{Hsieh2014,Inouye2014b,Inouye2015}.  More specifically we take the log likelihood of the node conditionals and add an $\ell_1$ penalty on all interaction terms:
\begin{align}
\argmin_{\multipall} -\sum_{\vi=1}^\vmax \left( \frac{1}{\imax} \sum_{\ii = 1}^\imax \left(\sum_{\j = 1}^\k \natp_{\j\vi\ii} \x_{\vi\ii}^{1/\j} - \A(\natpvec_{\vi\ii})\right) \right) + \lambda \sum_{\j=2}^{\k} \sum_{\ell=1}^{\j} \|\multip^{(\ell)}_{(\j)}\|_1 
\, ,
\end{align}
where $\natp_{\j\vi\ii} = \sum_{\ell=1}^\j  \left\langle \left[\multip^{(\ell)}_{(\j)}\right]_{\vi}, \ell \sqrt[\j]{\xvec_\ii} \,\circ^{\ell-1} \right\rangle$ and $\|\cdot\|_1$ is an entry-wise sum of absolute values.  Note that this is trivially decomposable into $\vmax$ subproblems and can thus be trivially parallelized to improve computation speed. We use the Newton-like method as in \cite{Hsieh2011,Inouye2015} to greatly reduce computation.  The initial innovation from \cite{Hsieh2011} was that the Hessian only needed to be computed over a \emph{free} set of variables each Newton iteration because of the $\ell_1$ regularization which suggested sparsity of the parameters.  Yet, the number of Newton iterations was very small compared to gradient descent.  In the case of GRM models, whose bottleneck is the computation of the gradient of $\A$ (at least under our current implementation though it might be possible to significantly reduce this bottleneck), this Newton-like method provides even more benefit because the gradient only has to be computed a small number of times (roughly 30) in our case rather than the several thousand times that would be needed for running thousands of proximal gradient descent steps for the same level of convergence.

In the next section, we derive the gradient and Hessian for the smooth part of the optimization as a function of the gradient and Hessian of the node conditional log partition function $\A(\natpvec)$.  Then, we develop a general method for bounding the log partition function $\A(\natpvec)$ and associated derivatives even though usually no closed-form exists.

\subsection{Gradient and Hessian of GRMs}
\paragraph{Notation for gradient and Hessian}
Let $\vectorize(\multip^{(\ell)}) \in \R^{\vmax^\ell}$ be the vectorized form of a tensor.  For example, the vectorized form of a $\vmax \times \vmax$ matrix is formed by stacking the matrix columns on top of each other to form one long $\vmax^2$ vector.  Also, let $[ x \given x \in \mathcal{X}]$ be analogous to the normal set notation $\{ x : x \in \mathcal{X} \}$ except that the bracket and vertical line notation creates a vector from all the elements concatenated to together.  This is similar to a list comprehension in Python.  For our gradient and Hessian calculations, we define the following variable transformations and give them as examples of this notation:
\newcommand{\zset}{\mathcal{Z}}
\newcommand{\regpset}{\mathcal{B}}
\begin{align*}
\regpset_\vi &= \big\{ \left[ \vectorize\left( \left[\multip^{(\ell)}_{(\j)}\right]_{\vi} \right) \midgiven \ell \leq \j \right] : \j \in \{1,2,\cdots,\k\}\big\} \\
&= \bigg\{
\underbrace{\left[\vectorize\left( \left[\multip^{(1)}_{(1)}\right]_{\vi} \right)\right]}_{\regpvec_{1\vi}},
\underbrace{\left[\vectorize\left( \left[\multip^{(1)}_{(2)}\right]_{\vi} \right),
\vectorize\left( \left[\multip^{(2)}_{(2)}\right]_{\vi} \right)\right]}_{\regpvec_{2\vi}},
\underbrace{\left[\vectorize\left( \left[\multip^{(1)}_{(3)}\right]_{\vi} \right), \cdots\right]}_{\regpvec_{3\vi}},
\cdots,
\underbrace{\left[ \cdots,
\vectorize\left( \left[\multip^{(\k)}_{(\k)}\right]_{\vi} \right)\right]}_{\regpvec_{\k\vi}}
\bigg\} , \\
\zset_{\vi\ii} &= \big\{ \left[ \vectorize\left( \ell \sqrt[\j]{\xvec_{\vi\ii}} \, \circ^{\ell-1} \right) \midgiven \ell \leq \j \right] : \j \in \{1,2,\cdots,\k\}\big\} \\
&= \bigg\{
\underbrace{\left[\vectorize\left( \sqrt[1]{\xvec_{\ii}} \, \circ^{0} \right)\right]}_{\zvec_{1\vi}},
\underbrace{\left[\vectorize\left( \sqrt[2]{\xvec_{\ii}} \, \circ^{0} \right),
\vectorize\left( 2\sqrt[2]{\xvec_{\ii}} \, \circ^{1} \right)\right]}_{\zvec_{2\vi}},
\underbrace{\left[\vectorize\left( \sqrt[3]{\xvec_{\ii}} \, \circ^{0} \right), \cdots\right]}_{\zvec_{3\vi}},
\cdots,
\underbrace{\left[ \cdots,
\vectorize\left( \k \sqrt[\k]{\xvec_{\ii}} \, \circ^{\k-1} \right)\right]}_{\zvec_{\k\vi}}
\bigg\} \, .
\end{align*}
With this notation, we have that $\natp_{\j\vi\ii} = \regp_{\j\vi}^T \zvec_{\j\vi\ii}$. Because each node regression is independent, we focus on solving one of the $\vmax$ subproblems for a particular $\vi$ using the notation from above:
\begin{align}
\argmin_{\regpset_\vi} \sum_{\ii=1}^\imax f_\vi(\regpset_\vi \given \x_{\vi\ii}, \zset_{\vi\ii} )\, ,
\end{align}
where $f_\vi(\regpset_\vi \given \x_{\vi\ii}, \zset_{\vi\ii} ) = -\sum_{\j=1}^\k (\regp_{\j\vi}^T \zvec_{\j\vi\ii}) \sqrt[\j]{\x_{\vi\ii}} + \A([\regp_{\j\vi}^T \zvec_{\j\vi\ii} \given \j \in \{1,\cdots,\k\})$.  For notational simplicity, we suppress the dependence on $\vi$ and $\ii$ in the derivations of the gradient and Hessian of $f(\cdot)$ (the gradient and Hessian are merely the sum over all instances).  With this simplified notation, the gradient and Hessian are as follows (as functions of $\A$, $\nabla \A$ and $\nabla^2 \A$):
\begin{align}
\nabla f(\regpset \given \x, \zset)
&= \left[\left(- \sqrt[\j]{\x} + \frac{\partial \A}{\partial \natp_{\j}} \right)\zvec_\j \midgiven \j \in \{1,2,\cdots,\k \} \right] \, ,\\
\nabla^2 f(\regpset \given \x, \zset) &= \left[
\begin{array}{l}
\left[\frac{\partial^2 \A}{\partial \natp_1 \partial\natp_\j} \zvec_\j \circ \zvec_1 \midgiven \j \in \{1,2,\cdots,\k\} \right], \vspace{0.5em} \\
\left[\frac{\partial^2 \A}{\partial \natp_2 \partial\natp_\j} \zvec_\j \circ \zvec_2 \midgiven \j \in \{1,2,\cdots,\k\}\right], \\
\hspace{5em} \vdots \\
\left[\frac{\partial^2 \A}{\partial \natp_\k \partial\natp_\j} \zvec_\j \circ \zvec_\k \midgiven \j \in \{1,2,\cdots,\k\}\right]
\end{array}
\right] \, .
\end{align}
Note how the gradient and Hessian are simple functions of $\zvec_\j$ and the derivatives of $\A(\natpvec)$. Thus, we develop bounded approximations for $\A(\natpvec)$, $\nabla\A(\natpvec)$ and $\nabla^2\A(\natpvec)$ next.

\subsection[Gradient and Hessian of A]{Gradient and Hessian of $\A(\natpvec)$}
\label{sec:gradient-hessian-A}
Because the node conditional distributions are not standard distributions, we must either derive the closed-form log partition function as done with the specific case of the exponential SQR model in \cite{Inouye2016}, or we must numerically approximate the log partition function and its first and second derivatives.  To the authors' best knowledge, even for the simplified SQR model with $\k=2$, no closed-form solution to log partition function exists for SQR node conditionals except for the discrete, Gaussian and exponential SQR models.  Thus, we seek a general way to estimate the log partition function and associated derivatives for any univariate exponential family; we also provide a concrete realization of this approximation method for the Poisson GRM case.

\paragraph[Derivatives of A Reformulated as Expectations]{Derivatives of $\A(\natpvec)$ Reformulated as Expectations}
We first note that the gradient and Hessian of $\A(\natpvec)$ are merely functions of particular expectations---a well-known result of exponential families:
\begin{align}
\A(\natpvec) &= \log \int_\domain \exp\Big(\sum_{\j=1}^\k \natp_\j \x^{\frac{1}{\j}} + \B(\x) \Big) \mathrm{d}\mu(\x) \\
\nabla \A(\natpvec) &= [ \E(\x^{\frac{1}{\j}}) \given \j \in \{1,\cdots,\k\} ] \\
\nabla^2 \A(\natpvec) &= \left[ 
\begin{array}{c}
\left[ \E(\x^{\frac{1}{\j}+\frac{1}{2}}) - \E(\x^{\frac{1}{\j}})\E(\x) \given \j \in \{1,\cdots,\k\} \right] \\
\left[ \E(\x^{\frac{1}{\j}+\frac{1}{2}}) - \E(\x^{\frac{1}{\j}})\E(\x^{\frac{1}{2}}) \given \j \in \{1,\cdots,\k\} \right] \\
\hspace{3em} \vdots \\
\left[ \E(\x^{\frac{1}{\j}+\frac{1}{\k}}) - \E(\x^{\frac{1}{\j}})\E(\x^{\frac{1}{\k}}) \given \j \in \{1,\cdots,\k\} \right] \\
\end{array}
\right]
\, .
\end{align}
Thus, we need to compute expectations for at most $\binom{\k}{2}+\k$ functions of the form $\E(\x^\a)$.

\paragraph{Definition of $\M(\a)$ to Unify Approximations}
To develop our approximations under a unified framework, let us define the following function $\M(\a)$ and its subfunctions denoted $f(\x)$ and $g(\x)$:
\begin{align}
\M(\a) &= \log \int_\domain \x^\a \exp\Big(\sum_{\j=1}^\k \natp_\j \x^{\frac{1}{\j}} + \B(\x) \Big) \mathrm{d}\mu(\x) = \log \int_\domain \exp\Big(\underbrace{\natp_1 \x + \B(\x)}_{f(\x)} + \underbrace{\textstyle{\sum_{\j=2}^\k}\, \natp_\j \x^{\frac{1}{\j}}  + \log(\x^\a)}_{g(\x)} \Big) \mathrm{d}\mu(\x) \, .
\label{eqn:M-function}
\end{align}
By simple inspection, we see that $\M(0) = \A(\natp_1,\natp_2)$ and $\E(\x^\a) = \exp\big( \M(\a)-\M(1) \big)$. Thus, by approximating $\M(\a)$, we can approximate all the necessary derivatives.  If $g(\x) = 0$, then this is simply the log partition function of the base exponential family, which is usually known in closed form.  If $g(\x) \approx b\x+c$ (as we will develop in the next sections), then we can create a modified $f$ and $g$ such that $\tilde{f}(\x) = (\natp_1+b)\x + c$ and $\tilde{g}(\x) = 0$---thus also allowing us to use the machinery of the base exponential family to compute the needed integrals.

\paragraph[Overall Approach to Bounding M(a)]{Overall Approach to Bounding $\M(\a)$}
Our approach splits the integral into $\nq = O(1)$ integrals which bound the integral over different subdomains of the domain.  We will choose the subdomains in appropriate way to minimize error, which will be described in a future section.  For each subdomain, we will form linear upper and lower bounds for $g(\x)$ so that we can then use the CDF function of the base exponential family to approximate the integrals over these subdomains.

First, we will describe how to compute linear upper and lower bounds to $g(\x)$ so that the integrals reduce to the original exponential family.  Because we can determine the concavity of each region of $g(\x)$,\footnote{This can be done by solving for the zeros of a polynomial.} we can form linear upper and lower bounds using the theory of convexity.  The secant line and the first-order Taylor series approximation form upper and lower bounds or vice versa depending on concavity. We can bound the tails of $g(\x)$ with a constant function or Taylor series approximation as appropriate.  See appendix for details on linear approximations for $g(\x)$.

If $g(\x)$ is upper and lower bounded by a linear functions, i.e. $b_l\x + c_l = g_l(\x) \leq g(\x) \leq g_u(\x) = b_u\x + c_u $, then we can form a modified functions of $f(\x)$ that will be upper and lower bounds of $f(\x)+g(\x)$:
\begin{align}
\begin{array}{r @{\,} c @{\,} l}
(\natp_1+b_l)\x + c_l = \hat{f}_l(\x) \leq & f(\x) + g(\x) & \leq \hat{f}_u(\x) = (\natp_1+b_u)\x + c_u 
\end{array}
.
\label{eqn:g-approx}
\end{align}
Assuming $\hat{\natp}_l = \natp_1 + b_l$ and $\hat{\natp}_u = \natp_1 + b_u$ are valid parameters, we can then use the original exponential family CDF---which is usually known in closed form---to compute the needed integrals.

Now that we have linear upper and lower bounds for $g(\x)$, we can upper and lower bound $\M(\a)$ using the CDF of the original exponential family to compute the needed integrals (see appendix for more derivation):
\begin{align}
\M(\a) &\approx \log \sum_{i=1}^\nq \int_{\domain_i} \exp(\hat{f}_i(\x)) \mathrm{d}\mu(\x) \\
&= \log \sum_{i=1}^\nq \exp\big(c_i + \A(\hat{\natp}_i) + \log\big(\text{CDF}\big(\max(\domain_i) \given \hat{\natp}_i\big) - \text{CDF}\big(\min(\domain_i) \given \hat{\natp}_i\big)\big) \, ,
\label{eqn:M-approx}
\end{align}
where the domain is split into disjoint subdomains, i.e. $\{ \domain_i : \domain = \bigcup_i^\nq \domain_i, \forall i \neq j, \domain_i \cap \domain_j = \emptyset \}$, $(\hat{\natp},b)$ are either $(\hat{\natp}_u,b_u)$ or $(\hat{\natp}_l,b_l)$ depending on whether the upper or lower bound is needed, $\A(\hat{\natp})$ and $\text{CDF}(\cdot)$ are the log partition function and CDF of the original exponential family.  Note that assuming $\A(\hat{\natp})$ and $\text{CDF}(\cdot)$ are available in closed form---as is the case for the Poisson distribution---this approximation can be computed in $O(\nq)=O(1)$ time.

\begin{figure}[!ht]
\centering
\includegraphics[width=0.4\textwidth, trim = 4cm 9.7cm 4cm 9.5cm, clip]{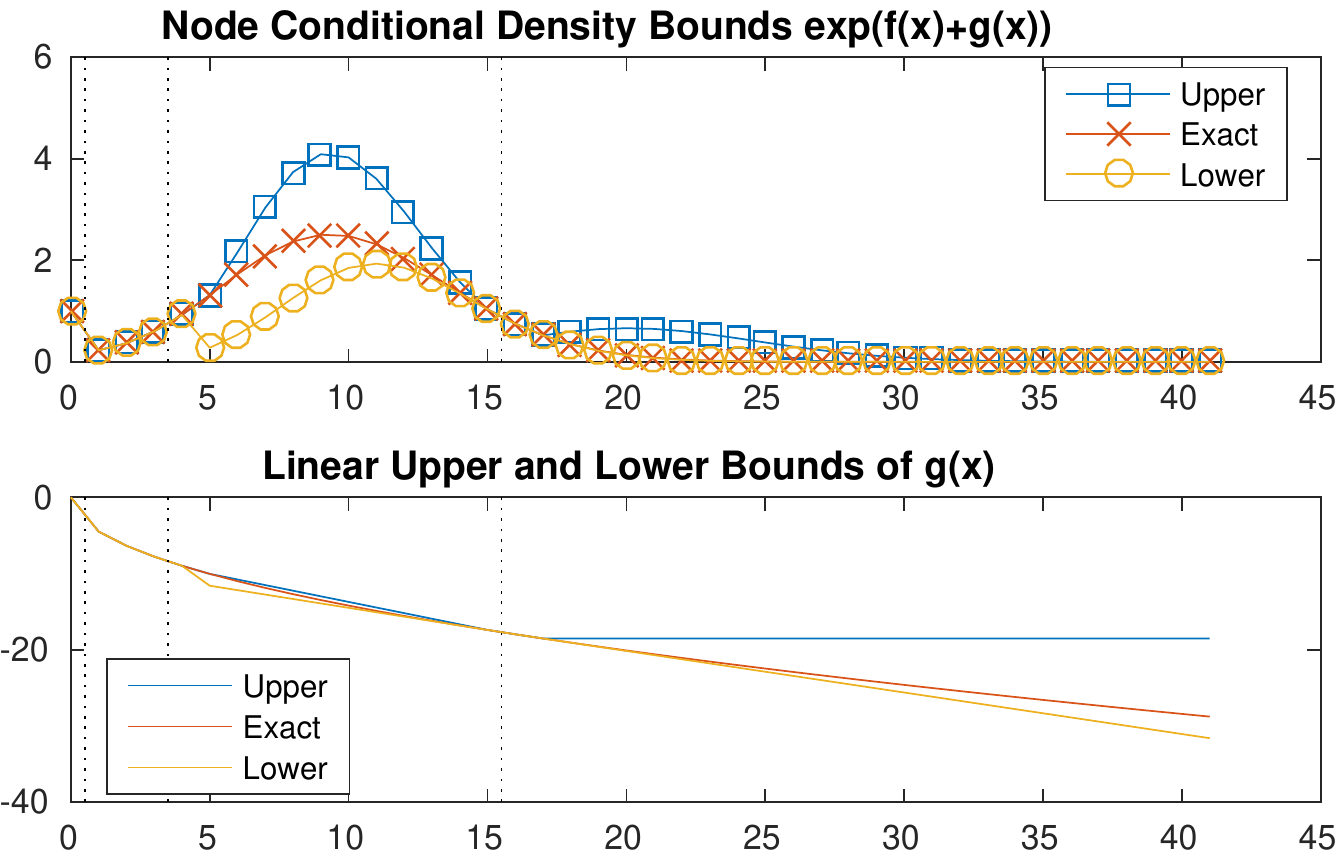}
\includegraphics[width=0.4\textwidth, trim = 4cm 9.7cm 4cm 9.5cm, clip]{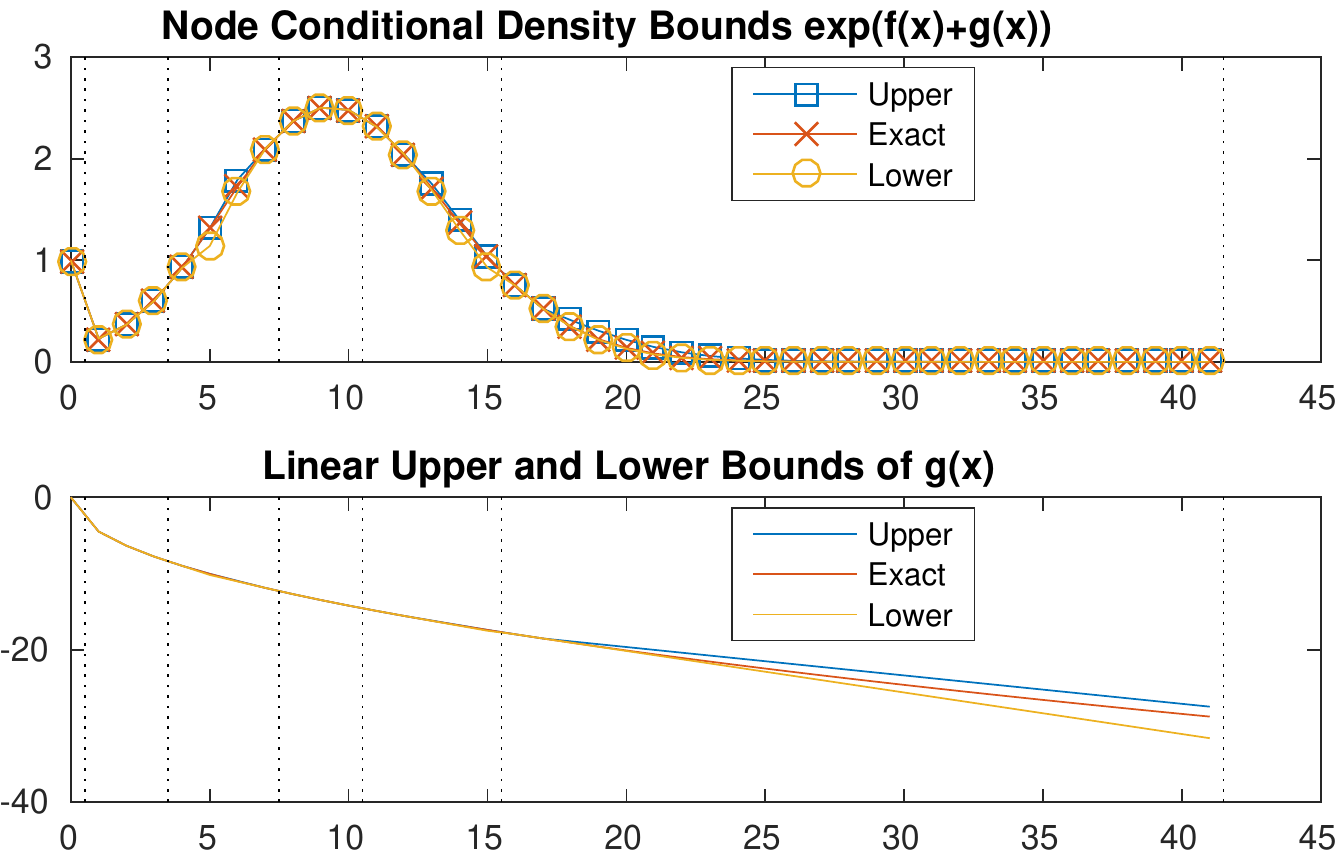}
\caption{Approximation of the $\M(\a)$ function with $a = 0$ and $\natpvec = [3.0232, -4.4966]$ for 2 subdomains (left) and for 5 subdomains (right) using the algorithm described in Sec.~\ref{sec:gradient-hessian-A}.  The top is the actual values of the summation in Eqn.~\ref{eqn:M-function} and the bottom is the linear approximation $b\x + c$ to the non-linear part $g(\x)$ as in Eqn.~\ref{eqn:g-approx}.  }
\label{fig:approx-illustration}
\end{figure}

\paragraph{Algorithm to Find Appropriate Subdomains $\domain_i$}
We need that every subdomain has a constant concavity (i.e. either concave or convex over the subdomain) in order to use Taylor series and secant line bounds (and a constant bound for the tails).  Thus, we use the following algorithm to find subdomains to help minimize the difference between the upper and lower bounds (An illustration of the method can be seen in Fig.~\ref{fig:approx-illustration}.):
\begin{enumerate}
\item Find all real roots of $g''(\x)$, denoted $(\x_1'',\x_2'',\cdots)$ so we know the inflection points (which will define the regions of constant concavity).
\item Use inflection points and endpoints of domain (e.g. $0$ and $\infty$ for Poisson) to define the initial subdomains.
\item Compute initial bounds for these subdomains using Eqn.~\ref{eqn:M-approx}.
\item Split the subdomain with the largest difference between upper and lower bounds (i.e. the subdomain with the largest error).
\item Recompute bounds for the two new subdomains formed by splitting the largest error subdomain.
\item Repeat previous two steps until $\nq$ domains have been obtained.
\end{enumerate}

Note that the roots of $g''(\x)$ can be solved by expanding to a polynomial and then computing the eigenvalues of the companion matrix.  For example if $g(\x) = \x^{1/2} + \x^{1/3} + \a\log(\x)$, then $g''(\x) = -\frac{1}{\x^2}(\frac{1}{4}\x^{1/2} + \frac{2}{9}\x^{1/3} + \a)$. Note that the zeros of this function are equal to the zeros of $h(x) =\frac{1}{4}\x^{1/2} + \frac{2}{9}\x^{1/3} + \a$.  Thus, we can let $y = \x^{1/6}$ and form the polynomial function $h(y) = \frac{1}{4}y^3 + \frac{2}{9}y^2 + \a$.  We can then solve the zeros of this polynomial by forming the companion matrix and solving for the eigenvalues.  However, we only need the real zeros and we do not care about multiplicity so it may be faster to use a direct root finding algorithm instead---though we have not explored this option.

\section{Results on Text Documents}
We computed the Poisson GRM model on two datasets: Classic3 and Grolier encyclopedia articles.  The Classic3 dataset contains 3893 research abstracts from library and information sciences, medical science and aeronautical engineering.  The Grolier encyclopedia dataset contains 5000 random articles from the Grolier encyclopedia.  We set $\k = 3$, $\vmax = 500$ and $\lambda = 0.01$ for our experiments. We chose 10 interval endpoints (i.e. 9 subdomains) for our approximations.  Note that this means there are at least $\binom{500}{3} \approx 2 \times 10^7$ possible parameters.  We give the top 10 positive parameters for individual, edge-wise and triple-wise combinations. The top 50 (unless there are less than 50 non-zeros) of both negative and positive dependencies for single, pairwise and triple-wise dependencies can be found in the appendix.
\begin{table}[!ht]
\caption{Table of Tuples}
\includegraphics[width=\textwidth, trim = 1.2cm 13.5cm 1.2cm 1.4cm, clip]{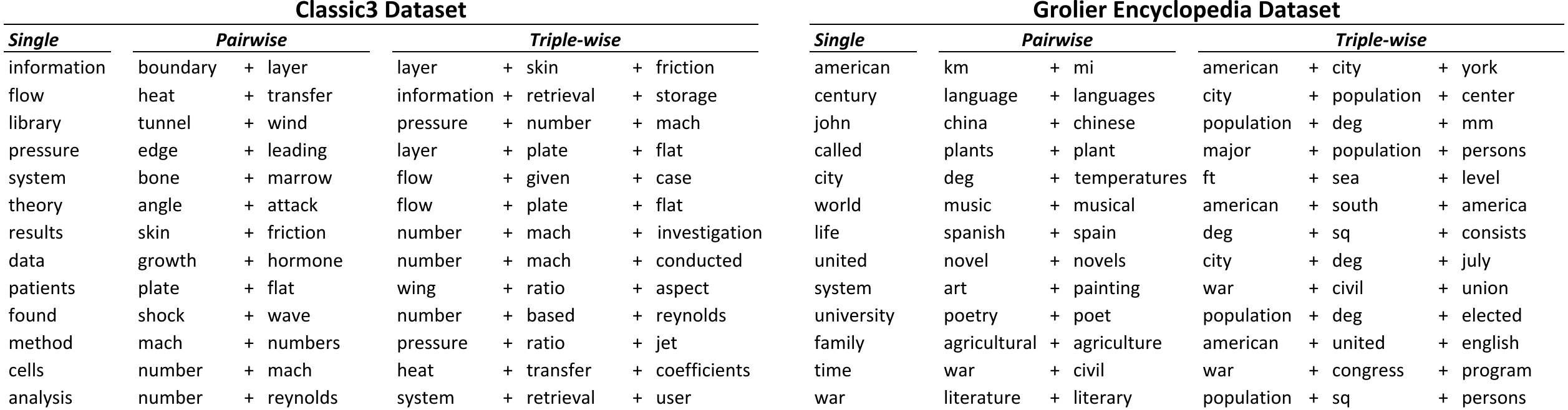}
\end{table}

These results illustrate that our model and algorithm can find interesting pairwise and triple-wise words.  The timing for these experiments using prototype code in MATLAB on TACC Maverick cluster (\url{https://portal.tacc.utexas.edu/user-guides/maverick}) was 2653 seconds for the Classic3 dataset and 5975 seconds for the Grolier dataset. Given the extremely large number of parameters to be optimized, this gives evidence that GRM models are computationally tractable while still wanting for some improvement.

\section{Discussion}
While it may seem at first that this model is impractical for even $\k = 4$, we suggest some practical ideas for reducing the parameter space.  First, if some parameters are known or expected a priori to be non-zero, we could only allow those parameters to be non-zero.  For example, known genetic pathways could be encoded as $\k$-wise cliques.  Thousands of known pathways could be added which would only incur thousands of parameters, which is very small relative to all possible parameters.  Second, the optimization could proceed in a stage-wise fashion such that the first a model is fit for $\k = 1$, then this model is used to choose which parameters to allow in the next model of $\k = 2$, etc.  For example, we could first train a model with only pairwise parameters ($\k =2$).  Then, we could find all triangles in the discovered graph and only add these parameters for training a model with $\k = 3$.  This heuristic would significantly reduce the number of possible parameters if the parameters are assumed to be sparse (as is usually the case with $\ell_1$-regularized objectives). Third, the tensors could be constrained to be low-rank and thus only $O(\vmax)$ values for each tensor would be needed.  For example, we could assume that the pairwise tensors are low-rank matrices.  For higher order tensors, a similar idea could hold, e.g. $\multip_{(\j)}^{(\ell)} = \sum_{i=1}^M \bm{\theta}_i \, \circ ^{\ell}$, where $M$ is $O(1)$.

\section{Conclusion}
We generalize the previous SQR \cite{Inouye2016} model to include factors of size $\k > 2$.  We study this general distribution by giving the node and radial conditional distributions, which provides simple conditions for normalization of the GRM class of models.  We then develop an approximation technique for estimating the node-wise log partition function and associated derivatives for the Poisson case---note that \cite{Inouye2016} only provided an algorithm for approximating the exponential SQR model.  Finally, we qualitatively demonstrated our model on two real world datasets.

\subsubsection*{Acknowledgments}
This work was supported by NSF (DGE-1110007, IIS-1149803, IIS-1447574, IIS-1546459, DMS-1264033, CCF-1320746) and ARO (W911NF-12-1-0390).

\small{
\bibliography{Mendeley-fixed}

\begin{thebibliography}{10}

\bibitem{Inouye2016}
D.~I. Inouye, P.~Ravikumar, and I.~S. Dhillon, ``Square root graphical models:
  Multivariate generalizations of univariate exponential families that permit
  positive dependencies,'' in {\em ICML}, 2016.

\bibitem{Ravikumar2010}
P.~Ravikumar, M.~Wainwright, and J.~Lafferty, ``High-dimensional {Ising} model
  selection using l1-regularized logistic regression,'' {\em The Annals of
  Statistics}, vol.~38, pp.~1287--1319, 6 2010.

\bibitem{Yang2015}
E.~Yang, P.~Ravikumar, G.~I. Allen, and Z.~Liu, ``On graphical models via
  univariate exponential family distributions,'' {\em JMLR}, vol.~16,
  pp.~3813--3847, 2015.

\bibitem{Inouye2015}
D.~I. Inouye, P.~Ravikumar, and I.~S. Dhillon, ``Fixed-length {Poisson} {MRF}:
  Adding dependencies to the multinomial,'' in {\em NIPS}, pp.~3195--3203,
  2015.

\bibitem{Hsieh2011}
C.~Hsieh, M.~Sustik, I.~Dhillon, and P.~Ravikumar, ``Sparse inverse covariance
  matrix estimation using quadratic approximation.,'' {\em Nips}, pp.~1--9,
  2011.

\bibitem{Yang2013}
E.~Yang, P.~Ravikumar, G.~I. Allen, and Z.~Liu, ``On {Poisson} graphical
  models,'' in {\em NIPS}, p.~1718–1726, 2013.

\bibitem{Blei2012}
D.~Blei, L.~Carin, and D.~Dunson, ``Probabilistic topic models,'' {\em IEEE
  Signal Processing Magazine}, vol.~27, pp.~55--65, 11 2010.

\bibitem{Blei2006}
D.~M.~B. Lafferty and J.~D., ``Correlated topic models,'' in {\em Advances in
  Neural Information Processing Systems 18}, pp.~147--154, 2006.

\bibitem{Inouye2014}
D.~I. Inouye, P.~Ravikumar, and I.~S. Dhillon, ``Admixture of {Poisson} {MRF}s:
  A topic model with word dependencies,'' in {\em ICML}, 2014.

\bibitem{Besag1974}
J.~Besag, ``Spatial interaction and the statistical analysis of lattice
  systems,'' {\em Journal of the Royal Statistical Society. Series B
  (Methodological)}, vol.~36, no.~2, pp.~192--236, 1974.

\bibitem{Hsieh2014}
C.-J. Hsieh, M.~A. Sustik, I.~S. Dhillon, and P.~Ravikumar, ``{QUIC}: Quadratic
  approximation for sparse inverse covariance estimation,'' {\em JMLR},
  vol.~15, pp.~2911--2947, 2014.

\bibitem{Inouye2014b}
D.~I. Inouye, P.~Ravikumar, and I.~S. Dhillon, ``Capturing semantically
  meaningful word dependencies with an admixture of {\{}{poisson} {mrf}{\}s},''
  in {\em NIPS}, pp.~3158--3166, 2014.

\end{thebibliography}
\bibliographystyle{ieeetr}
}

\clearpage
\appendix

\section{Node Conditional Derivation}
\begin{align}
\Pr(\x_\vi \given \xvec_{-\vi}, \multipall) &= \Pr(\xvec = \xvec_{\vi0} + \x_\vi \evec_\vi \given \xvec_{\vi0}, \multipall) \\
&\propto \exp\left( \sum_{\j=1}^\k \sum_{\ell=1}^\j \langle \multip^{(\ell)}_{(\j)}, \sqrt[\j]{\xvec_{\vi0} + \x_\vi\evec_\vi} \, \circ^{\ell} \rangle + \B(\x_\vi) \right) \\
&\propto \exp\left( \sum_{\j=1}^\k \sum_{\ell=1}^\j \langle \multip^{(\ell)}_{(\j)}, (\sqrt[\j]{\xvec_{\vi0}} + \sqrt[\j]{\x_\vi\evec_\vi}) \, \circ^{\ell} \rangle + \B(\x_\vi) \right) \\
&\propto \exp\left( \sum_{\j=1}^\k \sum_{\ell=1}^\j \left\langle \multip^{(\ell)}_{(\j)}, \sum_{m=0}^\ell \binom{\ell}{m} \left(\sqrt[\j]{\xvec_{\vi0}} \,\circ^{\ell-m}\right) \circ \left(\sqrt[\j]{\x_\vi\evec_\vi} \, \circ^{m}\right) \right\rangle + \B(\x_\vi) \right) \\
&\propto \exp\left( \sum_{\j=1}^\k \sum_{\ell=1}^\j  \sum_{m=0}^\ell \binom{\ell}{m}  \left\langle \multip^{(\ell)}_{(\j)},\left(\sqrt[\j]{\xvec_{\vi0}} \,\circ^{\ell-m}\right) \circ \left(\sqrt[\j]{\x_\vi\evec_\vi} \, \circ^{m}\right) \right\rangle + \B(\x_\vi) \right) \\
&\propto \exp\left( \sum_{\j=1}^\k \sum_{\ell=1}^\j  \sum_{m=0}^\ell \binom{\ell}{m} \left\langle \left[\multip^{(\ell)}_{(\j)}\right]_{\mathbf{I}(\vi,m)},\sqrt[\j]{\xvec_{\vi0}} \,\circ^{\ell-m} \right\rangle \x_\vi^{m/\j} + \B(\x_\vi) \right) \\
&\propto \exp\left( \sum_{\j=1}^\k \sum_{\ell=1}^\j  \sum_{m=1}^\ell \binom{\ell}{m} \left\langle \left[\multip^{(\ell)}_{(\j)}\right]_{\mathbf{I}(\vi,m)},\sqrt[\j]{\xvec_{\vi0}} \,\circ^{\ell-m} \right\rangle \x_\vi^{m/\j} + \B(\x_\vi) \right) \tag{$m=0$ is constant}\\
&\propto \exp\left( \sum_{\j=1}^\k \sum_{\ell=1}^\j  \binom{\ell}{1} \left\langle \left[\multip^{(\ell)}_{(\j)}\right]_{\mathbf{I}(\vi,1)},\sqrt[\j]{\xvec_{\vi0}} \,\circ^{\ell-1} \right\rangle \x_\vi^{1/\j} + \B(\x_\vi) \right) \tag{$m\geq 2$ are all zero since subtensors are zero by construction}\\
&\propto \exp\left( \sum_{\j=1}^\k \left( \sum_{\ell=1}^\j  \ell \left\langle \left[\multip^{(\ell)}_{(\j)}\right]_{\mathbf{I}(\vi,1)},\sqrt[\j]{\xvec_{\vi0}} \,\circ^{\ell-1} \right\rangle \right) \x_\vi^{1/\j} + \B(\x_\vi) \right) \\
&\propto \exp\left( \sum_{\j=1}^\k \natp_{\j\vi} \x_\vi^{1/\j} + \B(\x_\vi) \right)
\, ,
\end{align}
where $\natp_{\j\vi} = \left( \sum_{\ell=1}^\j  \ell \left\langle \left[\multip^{(\ell)}_{(\j)}\right]_{\vi},\sqrt[\j]{\xvec_{\vi0}} \,\circ^{\ell-1} \right\rangle \right)$.  See notation section for definition of $\left[\multip^{(\ell)}_{(\j)}\right]_{\vi}$.  This is a univariate exponential family with sufficient statistics $\x_\vi^{1/\j}$, natural parameters $\natp_{\j\vi}$, and base measure $\B(\x_\vi)$. This recovers the SQR node conditional from \cite{Inouye2016} with $\k = 2$.  

\section{Radial Conditional Derivation}
As in \cite{Inouye2016}, we define the \emph{radial} conditional distribution by fixing the unit direction $\uvec = \frac{\xvec}{\|\xvec\|_1}$ of the sufficient statistics but allowing the scaling $\radx = \|\xvec\|_1$ to be unkown.  Thus, we get the following \emph{radial} conditional distribution:
\begin{align}
\Pr( \instvec = \radx \uvec \given \uvec, \multipall ) &\propto \exp\left( \sum_{\j=1}^\k \sum_{\ell=1}^\j \langle \multip^{(\ell)}_{(\j)}, \sqrt[\j]{\radx\uvec} \, \circ^{\ell} \rangle + \textstyle{\sum_{\vi}} \B(\radx\usca_\vi) \right) \\
&\propto \exp\left( \sum_{\j=1}^\k \sum_{\ell=1}^\j \langle \multip^{(\ell)}_{(\j)}, \sqrt[\j]{\uvec} \, \circ^{\ell} \rangle \radx^{\frac{\ell}{\j}} + \textstyle{\sum_{\vi}} \B(\radx\usca_\vi) \right) \\
&\propto \exp\left( \sum_{r\in\mathcal{R}} \natp_r(\uvec) \radx^r + \tilde{\B}_{\uvec}(\radx) \right)
\, ,
\end{align}
where $\mathcal{R} = \{\ell/\j : \j \in \{1,\cdots,\k\}, \ell \in \{1,\cdots,\j\}\}$ is the set of possible ratios, $\natp_r(\uvec) = \sum_{\{(\ell,\j): \ell/\j = r\} } \langle \multip^{(\ell)}_{(\j)}, \sqrt[\j]{\uvec} \, \circ^{\ell} \rangle$ are the exponential family parameters, $\radx^r$ are the corresponding sufficient statistics, and $\tilde{\B}_{\uvec}(\radx) = \textstyle{\sum_{\vi}} \B(\radx\usca_\vi)$ is the base measure.  Thus, the radial conditional distribution is a univariate exponential family.

\section[Derivation of M(a)]{Derivation of $\M(\a)$ Approximation}
\begin{align}
\M(\a) &\approx \log \sum_{i=1}^\nq \int_{\domain_i} \exp(\hat{f}_i(\x)) \mathrm{d}\mu(\x) \\
&= \log \sum_{i=1}^\nq \exp(c_i) \int_{\domain_i} \exp(\hat{\natp}_i \x + \B(\x)) \mathrm{d}\mu(\x) \\
&= \log \sum_{i=1}^\nq \exp(c_i)\exp\big(\A(\hat{\natp}_i)\big)\Big(\text{CDF}\big(\max(\domain_i) \given \hat{\natp}_i\big) - \text{CDF}\big(\min(\domain_i) \given \hat{\natp}_i\big)\Big) \\
&= \log \sum_{i=1}^\nq \exp\big(c_i + \A(\hat{\natp}_i)\big)\Big(\text{CDF}\big(\max(\domain_i) \given \hat{\natp}_i\big) - \text{CDF}\big(\min(\domain_i) \given \hat{\natp}_i\big)\Big) \\
&= \log \sum_{i=1}^\nq \exp\big(c_i + \A(\hat{\natp}_i) + \log\big(\text{CDF}\big(\max(\domain_i) \given \hat{\natp}_i\big) - \text{CDF}\big(\min(\domain_i) \given \hat{\natp}_i\big)\big) \, ,
\end{align}

\section[Linear Bounds of g(x)]{Linear Bounds of $g(\x)$}
\paragraph{Taylor series linear bound} Upper bound if concavity = -1 and lower bound if concavity = 1:
\begin{align*}
\q^* &= \left\{ \begin{array}{ll}
\q_1 &, \text{if}\, \q_2 = \infty \\
\q_2 &, \text{if}\, \q_1 = -\infty \\
\argmax_{\q_1,\q_2} g(\q) &, \text{otherwise}
\end{array} \right\} \\
\hat{g}(\x) &= g(\q^*) + g'(\q^*)(\x-\q^*) \\
&= \underbrace{g'(\q^*)}_{b} \x + \underbrace{\big(g(\q^*) - \q^* g'(\q^*)\big)}_{c}
\end{align*}

\paragraph{Secant linear bound} Upper bound if concavity = 1 and lower bound if concavity = -1:
\begin{align*}
b &=\frac{g(\q_2)-g(\q_1)}{\q_2-\q_1} \\
g(\q_1) &= b\q_1 + c \\
\Rightarrow c &= g(\q_1) - b\q_1
\end{align*}

\paragraph{Tail bounds} We know there are only a finite number of inflection points so let us take the $\x$ value for the last inflection point, denoted $\x^*$.  By simple asymptotic analysis, we know that the largest non-zero term will dominate eventually.  Let's assume w.l.o.g. that $\natp_{\j^*} \x^{\frac{1}{\j^*}}$ dominates\footnote{If for all $\j$, $\natp_\j = 0$, then we can take $\j^* = \infty$, $\natp_{\j^*} = \a$.} and $\natp_{\j^*} > 0$.  Then, we know that after the last inflection point, the concavity will be negative.  In addition, we know that the $g(\x) \to \infty$ as $\x \to \infty$.  The function must be monotonically increasing after the last inflection point.  Proof by contradiction: Suppose the monotonicity is negative after the last inflection point. Then, because the $g(\x)$ is a continuous function and $g(\x) \to \infty$ as $\x \to \infty$, the function must eventually have a positive monotonicity.  Yet this would switch from negative monotonicity to positive monotonicity after the last inflection point.  However, this would be an inflection point that is greater than the assumed last inflection point which leads to a contradiction.  The case where $\natp_{\j*} < 0$ can be proved similarly.  Thus, we can use a constant function for an upper bound if concavity = 1. and we can use a constant function as a lower bound if concavity = -1.  A Taylor series approximation forms an upper or lower bound depending on concavity.

\section{Complete Results for Classic3 Dataset}
\begin{verbatim}
<<< Largest 1-tuples, j = 1, ell = 1 >>>
-0.6256  information
-0.6906  flow
-0.8097  library
-1.1693  pressure
-1.4066  system
-1.4090  theory
-1.4209  results
-1.4248  data
-1.4597  patients
-1.5737  found
-1.5791  method
-1.6292  cells
-1.6566  analysis
-1.7161  given
-1.7198  use
-1.7389  number
-1.7525  used
-1.7550  study
-1.7860  made
-1.7884  effect
-1.8054  time
-1.8089  body
-1.8371  research
-1.8563  cases
-1.8958  normal
-1.9246  effects
-1.9436  present
-1.9690  discussed
-1.9803  shock
-1.9899  presented
-2.0112  wing
-2.0135  surface
-2.0275  large
-2.0295  case
-2.0447  obtained
-2.0603  new
-2.0695  paper
-2.0776  libraries
-2.0834  high
-2.0884  problems
-2.1160  methods
-2.1162  well
-2.1311  development
-2.1414  general
-2.1417  growth
-2.1659  problem
-2.2109  jet
-2.2112  terms
-2.2313  systems
-2.2416  form

<<< Positive 2-tuples, j = 2, ell = 2 >>>
 4.9827  boundary + layer
 4.2583  heat + transfer
 3.9493  tunnel + wind
 3.3190  edge + leading
 3.1510  bone + marrow
 3.0395  angle + attack
 2.8711  skin + friction
 2.5638  growth + hormone
 2.3182  plate + flat
 2.2675  shock + wave
 2.2548  mach + numbers
 2.1277  number + mach
 2.1047  number + reynolds
 2.0561  agreement + good
 2.0306  attack + angles
 2.0102  document + documents
 1.9377  cells + cell
 1.7727  journals + journal
 1.5436  library + libraries
 1.5291  lift + drag
 1.5090  wing + wings
 1.4283  shells + cylindrical
 1.4192  buckling + shells
 1.4126  temperature + thermal
 1.4080  free + stream
 1.3980  ratio + aspect
 1.3892  equations + differential
 1.3721  boundary + layers
 1.3675  point + stagnation
 1.2517  shock + waves
 1.2495  heat + temperature
 1.2285  reynolds + transition
 1.2000  wings + aspect
 1.1793  temperature + temperatures
 1.1554  thin + shells
 1.1335  science + scientific
 1.1279  cells + marrow
 1.1188  numbers + reynolds
 1.1004  cylinder + circular
 1.0910  renal + kidney
 1.0907  pressure + pressures
 1.0384  high + speed
 1.0294  layer + laminar
 1.0266  information + retrieval
 1.0232  patients + therapy
 1.0159  patients + cancer
 1.0127  jet + nozzle
 0.9848  group + groups
 0.9660  experimental + theoretical
 0.9528  buckling + stress

<<< Negative 2-tuples, j = 2, ell = 2 >>>
-0.8428  flow - library
-0.6400  information - pressure
-0.5896  flow - cells
-0.5885  pressure - library
-0.5737  library - patients
-0.5570  flow - system
-0.5559  information - cells
-0.5533  information - heat
-0.5185  information - patients
-0.5184  flow - patients
-0.4941  theory - patients
-0.4902  information - normal
-0.4695  information - found
-0.4600  information - effect
-0.4587  library - theory
-0.4489  library - normal
-0.4479  library - cells
-0.4213  library - effects
-0.4097  flow - retrieval
-0.4071  flow - growth
-0.3925  library - found
-0.3857  library - cases
-0.3667  information - wing
-0.3653  information - case
-0.3580  pressure - cells
-0.3548  flow - information
-0.3383  flow - subject
-0.3364  results - library
-0.3316  information - effects
-0.3242  information - temperature
-0.3170  information - surface
-0.3143  flow - children
-0.3131  library - obtained
-0.2957  flow - book
-0.2904  flow - research
-0.2903  information - mach
-0.2863  theory - cells
-0.2858  library - effect
-0.2824  information - equations
-0.2817  flow - literature
-0.2780  flow - index
-0.2754  flow - buckling
-0.2745  analysis - patients
-0.2692  information - cases
-0.2650  information - shock
-0.2626  information - boundary
-0.2583  information - method
-0.2566  information - high
-0.2524  library - body
-0.2522  information - ratio

<<< Positive 3-tuples, j = 3, ell = 3 >>>
 0.5067  layer + skin + friction
 0.3171  information + retrieval + storage
 0.3149  pressure + number + mach
 0.3118  layer + plate + flat
 0.2672  flow + given + case
 0.2411  flow + plate + flat
 0.1759  number + mach + investigation
 0.1390  number + mach + conducted
 0.1340  wing + ratio + aspect
 0.1317  number + based + reynolds
 0.1100  pressure + ratio + jet
 0.1072  heat + transfer + coefficients
 0.0973  system + retrieval + user
 0.0972  boundary + layer + experiments
 0.0926  mach + free + stream
 0.0862  pressure + layer + gradient
 0.0825  heat + temperature + coefficient
 0.0716  pressure + supersonic + base
 0.0711  boundary + shock + interaction
 0.0709  boundary + layer + distance
 0.0709  layer + shock + interaction
 0.0678  theory + experimental + experiment
 0.0594  flow + fluid + steady
 0.0578  flow + boundary + present
 0.0554  flow + body + revolution
 0.0537  flow + case + form
 0.0524  flow + body + shape
 0.0511  information + data + base
 0.0479  boundary + layer + found
 0.0478  cells + bone + marrow
 0.0462  flow + theory + approximation
 0.0457  data + retrieval + base
 0.0430  results + number + higher
 0.0420  layer + temperature + compressible
 0.0417  number + mach + static
 0.0403  boundary + injection + mass
 0.0397  number + mach + approximately
 0.0393  flow + hypersonic + region
 0.0365  theory + wing + wings
 0.0360  growth + human + hormone
 0.0359  number + mach + lower
 0.0357  heat + transfer + blunt
 0.0337  number + mach + increasing
 0.0337  number + boundary + increasing
 0.0328  boundary + layer + measurements
 0.0306  number + boundary + reynolds
 0.0305  flow + body + conditions
 0.0300  information + field + science
 0.0293  flow + number + based
 0.0290  flow + data + experimental

<<< Negative 3-tuples, j = 3, ell = 3 >>>
-0.3490  boundary - layer - conditions
-0.2025  number - mach - numbers
-0.0907  boundary - layer - wing
-0.0566  flow - number - numbers
-0.0548  boundary - layer - time
-0.0450  layer - shock - laminar
-0.0433  number - mach - solution
-0.0353  boundary - layer - jet
-0.0274  heat - transfer - jet
-0.0265  boundary - solutions - turbulent
-0.0236  flow - mach - reynolds
-0.0208  pressure - number - numbers
-0.0101  boundary - layer - flutter
-0.0019  flow - mach - velocity
-0.0014  number - mach - problems
\end{verbatim}

\section{Complete Results for Grolier Encyclopedia Dataset}
\begin{verbatim}
<<< Largest 1-tuples, j = 1, ell = 1 >>>
-1.6202  american
-1.7936  century
-1.8188  john
-1.8830  called
-1.8866  city
-1.9162  world
-1.9543  life
-2.0359  united
-2.1299  system
-2.1328  university
-2.1390  family
-2.1473  time
-2.1591  war
-2.1858  include
-2.1870  english
-2.2457  water
-2.2485  history
-2.2559  de
-2.2694  form
-2.3326  major
-2.3442  national
-2.3523  french
-2.3537  william
-2.3708  art
-2.3808  found
-2.4045  name
-2.4049  modern
-2.4255  music
-2.4315  power
-2.4433  king
-2.4445  social
-2.4455  british
-2.4596  usually
-2.4718  charles
-2.4784  south
-2.4923  law
-2.4995  north
-2.5030  repr
-2.5165  species
-2.5247  theory
-2.5383  human
-2.5520  ft
-2.5530  black
-2.5566  government
-2.5660  west
-2.5773  york
-2.5777  church
-2.5841  school
-2.5890  development
-2.5899  common

<<< Positive 2-tuples, j = 2, ell = 2 >>>
 8.7140  km + mi
 3.9800  language + languages
 2.9617  china + chinese
 2.6237  plants + plant
 2.5229  deg + temperatures
 2.5152  music + musical
 2.4147  spanish + spain
 2.1495  novel + novels
 2.1059  art + painting
 2.0869  poetry + poet
 2.0738  agricultural + agriculture
 2.0492  war + civil
 2.0024  literature + literary
 1.8595  french + france
 1.8405  german + germany
 1.7779  culture + cultural
 1.7453  china + asia
 1.7368  india + asia
 1.7088  system + systems
 1.6836  city + york
 1.6818  west + east
 1.5932  africa + african
 1.5932  deg + mm
 1.5137  southern + northern
 1.4987  architecture + building
 1.4788  style + architecture
 1.4431  body + blood
 1.4359  role + played
 1.4313  sea + ocean
 1.4304  cells + blood
 1.4118  science + scientific
 1.4074  century + centuries
 1.3808  population + sq
 1.3723  social + society
 1.3611  italian + renaissance
 1.3611  music + opera
 1.3600  ocean + pacific
 1.3569  cause + disease
 1.3548  cities + urban
 1.3419  war + army
 1.3308  united + countries
 1.3065  animals + animal
 1.2998  church + christian
 1.2991  art + museum
 1.2907  education + schools
 1.2809  programs + program
 1.2715  deg + temperature
 1.2550  world + war
 1.2515  party + leader
 1.2477  government + federal

<<< Negative 2-tuples, j = 2, ell = 2 >>>
-0.2449  life - languages
-0.2179  century - species
-0.2041  city - species
-0.1575  war - species
-0.1156  city - sq
-0.0941  century - june
-0.0911  war - languages
-0.0771  war - example
-0.0725  city - theory
-0.0718  city - common
-0.0707  city - system
-0.0684  city - called
-0.0666  century - cells
-0.0623  war - cells
-0.0501  american - eng
-0.0436  city - english
-0.0420  century - president
-0.0415  american - ft
-0.0379  art - america
-0.0367  city - found
-0.0359  city - form
-0.0340  city - development
-0.0338  war - form
-0.0333  war - usually
-0.0301  called - deg
-0.0279  war - forms
-0.0247  city - time
-0.0241  city - united
-0.0237  war - human
-0.0199  century - july
-0.0199  century - party
-0.0152  war - theory
-0.0151  american - king
-0.0128  city - family
-0.0119  century - south
-0.0112  called - eng
-0.0109  city - life
-0.0103  american - deg
-0.0102  american - east
-0.0080  american - city
-0.0071  war - water
-0.0058  city - usually
-0.0053  american - cells
-0.0051  form - university
-0.0046  city - body
-0.0039  city - process
-0.0029  century - american
-0.0027  ft - english
-0.0008  city - cells

<<< Positive 3-tuples, j = 3, ell = 3 >>>
 0.3126  american + city + york
 0.2773  city + population + center
 0.2549  population + deg + mm
 0.1971  major + population + persons
 0.1641  ft + sea + level
 0.1523  american + south + america
 0.1515  deg + sq + consists
 0.1351  city + deg + july
 0.1330  war + civil + union
 0.1173  population + deg + elected
 0.1170  american + united + english
 0.1140  war + congress + program
 0.1054  population + sq + persons
 0.0986  american + french + british
 0.0972  language + includes + languages
 0.0965  world + war + japanese
 0.0923  major + time + changes
 0.0899  century + world + laws
 0.0879  life + human + stage
 0.0875  north + south + president
 0.0870  city + population + university
 0.0845  century + world + war
 0.0830  km + mi + discovered
 0.0813  war + united + received
 0.0770  city + river + historical
 0.0760  century + history + short
 0.0744  century + english + story
 0.0738  war + south + union
 0.0721  american + war + congress
 0.0698  world + united + david
 0.0676  century + history + active
 0.0674  century + history + wide
 0.0669  war + united + america
 0.0668  war + army + june
 0.0664  city + km + population
 0.0664  government + national + rise
 0.0663  century + form + appeared
 0.0649  war + united + caused
 0.0644  century + world + separate
 0.0642  united + people + continued
 0.0638  city + united + urban
 0.0638  century + time + applied
 0.0633  world + united + building
 0.0629  world + south + iron
 0.0621  world + population + rate
 0.0614  city + university + center
 0.0605  century + time + studied
 0.0599  american + war + people
 0.0586  north + km + fish
 0.0582  world + william + series

<<< Negative 3-tuples, j = 3, ell = 3 >>>
-0.2560  city - population - york
-0.1206  km - mi - america
-0.1076  american - km - mi
-0.0586  km - mi - york
-0.0518  war - north - example
-0.0429  km - mi - social
-0.0413  km - mi - family
-0.0294  km - mi - own
-0.0293  km - mi - style
-0.0263  city - population - style
-0.0262  km - mi - theory
-0.0253  city - center - sq
-0.0208  km - mi - law
-0.0207  km - mi - human
-0.0182  km - mi - example
-0.0157  city - population - greek
-0.0145  km - mi - water
-0.0125  called - km - mi
-0.0114  england - language - languages
-0.0099  km - mi - greek
-0.0093  time - km - mi
-0.0042  km - mi - english
-0.0025  mi - population - america
-0.0021  population - america - sq
-0.0012  war - km - mi
-0.0003  american - city - center
\end{verbatim}

\end{document}